# Experience-based Refinement of Task Planning Knowledge in Autonomous Robots

Hadeel Jazzaa[a], Thomas McCluskey[a] and David Peebles[b]

[a]School of Computing and Engineering, University of Huddersfield, Huddersfield, UK
[b]School of Human and Health Sciences, University of Huddersfield, Huddersfield, UK

ABSTRACT

The requirement for autonomous robots to exhibit higher-level cognitive skills by planning and adapting in an ever changing environment is indeed a great challenge for the AI community. Progress has been made in the automated planning community on refinement and repair of an agent's symbolic knowledge to do task planning in an incomplete or changing environmental model, but these advances up to now have not been transferred to real, physical robots. This paper demonstrates how a physical robot can be capable of adapting its symbolic knowledge of the environment, by using experiences in robot action execution to drive knowledge refinement and hence to improve the success rate of the task plans the robot creates. To implement more robust planning systems, we propose a method for refining domain knowledge to improve the knowledge on which intelligent robot behaviour is based. This architecture has been implemented and evaluated using a NAO robot. The refined knowledge leads to the future synthesis of task plans which demonstrate decreasing rates of failure over time as faulty knowledge is removed or adjusted.

## 1. Introduction

Autonomous robots operating in partially known and dynamic environments (e.g. exploration robotics) require systems to deliberately create and execute plans [21]. Gathering knowledge and acting on that knowledge, in such environments is a major demand for robust autonomous robot systems.

A robot may encounter a range of failures while executing its planned actions. The plan execution strategy must account for the action/plan failure, which results from ignorance or change [7]. The robot could overcome such failures by gathering updated information about its current environment, and re-planning to generate a new plan. However, in some cases re-planning does not solve the problem because the robot's knowledge of the conditions and effects of its own actions on the environment may be faulty.

Progress has been made in the automated planning community on refinement and repair of an agent's symbolic knowledge to do task planning in an incomplete or changing environmental model, but most contributions focus on enhancing the efficiency of plan generation within task planners. More seriously, the published methods are invariably tied to benchmarks and virtual environments rather than real physical worlds. Only a limited number of studies address the aspect of learning while acting, and even fewer have demonstrated applications in robotics [22]. Furthermore, integrated representation between planning and acting is a weak point in most systems, except for a few that have state variables hierarchy and continuity between acting and planning [21].

hadeel.jazzaa@hud.ac.uk (H. Jazzaa); t.l.mccluskey@hud.ac.uk (T. McCluskey); d.peebles@hud.ac.uk (D. Peebles)
ORCID(s): 0009-0000-3181-411X (H. Jazzaa); 0000-0001-8181-8127 (T. McCluskey); 0000-0003-1008-9275 (D. Peebles)

Equipping autonomous robots with reasoning mechanisms can be useful when dealing with such issues. To overcome failure, the robot must reason about the cause of the failure to refine its knowledge using its experience. When execution of an action in a plan fails, the robot must realise why it has happened. Then, to avoid any future failure, an update to its knowledge should be made by the refinement system.

Systems with pre-engineered knowledge of their environment often provide an abstraction hierarchy of types of knowledge for domain experts to use in crafting knowledge. At an abstract level parameterised actions are key to an efficient approach in bridging the gap between the abstract plan and action execution. Actions include parameters that accept changed values to enable the implementation of the same action in different situations [40, 4]. For example, the action *GoTo(Pos)* is more flexible to program than *GoToCentre*, where *Pos* is a parameter that accepts different values, including the centre [40].

But however much hand-crafted engineering is facilitated, incomplete action models that are missing some aspects may cause action execution failure and result in failures in achieving tasks. Hence, ultimately, plan construction and execution can only be successful through a *combination* of human-specified and robot-learned knowledge [40] on which behaviour is based. For example, incorrectly specified actions requiring new attributes or relationships, in the face of a changing environment needs to be updated regularly without relying always on human hand-crafting. Additionally, an engineer may miss preconditions, effects, correct knowledge, or entire action definitions may be absent from the model [15].

Our work furthers the hypothesis that accurate models can be learned online while data gathering during operation

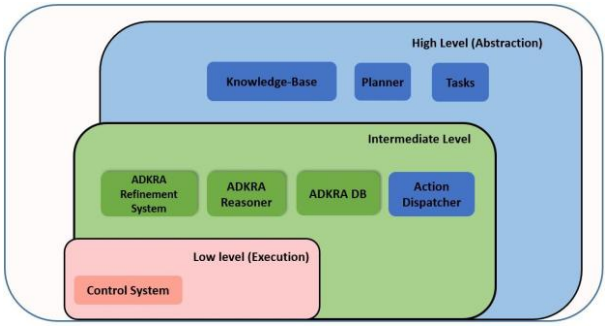

**Figure 1:** HDJ Intelligent Robot Hierarchical Architecture. The existing components (Blue boxes) and ADKRA components (Green boxes)

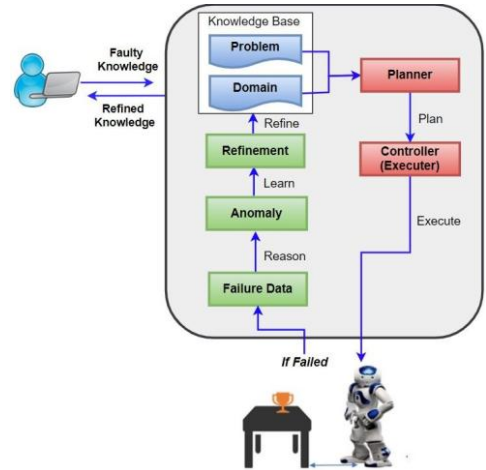

**Figure 2:** Conceptual Model for Refining Task Planning Domain Knowledge

[33]. This is aligned with the move towards long-term autonomy [27], in that making the system more robust through experiential learning will increase its effective life.

In this paper we introduce a general approach for the automated knowledge refinement of planning domain models used in task planning. It drives the improvement of pre-specified domain knowledge to adapt to changing environments, reducing the reliance on manual craftsmanship and addressing the refinement of incorrectly specified parameterised actions by human engineers. We specify a means to utilise this refinement approach within a real, physical robot architecture, and through experience-based learning during execution improve its domain knowledge so that future task plans are less likely to fail due to incorrect assumptions about the world. Finally, we demonstrate an empirical evaluation of the robot architecture and knowledge refinement capabilities using a real-world humanoid robot within a kitchen scenario.

While robot architectures have been equipped with various types of learning, and there are task planning methods that can learn to repair faulty knowledge by experience, as far as we are aware this is the first work to combine these together for the benefit of task planning in robots: we have designed, implemented and evaluated a method in which a physical robot can repair its high level knowledge by experience.

The paper is structured as follows: Section 2 provides an overview of the method we propose. Section 3 provides an in-depth description of our proposed robotics hierarchical architecture (called HDJ after the first author), including its framework, system mechanism, and algorithms. Section 4 details the implementation of HDJ using the domain and case study. Section 5 focuses on the evaluation of the HDJ system, describing the design and execution of a series of experiments. Section 6 offers an overview of related research work in the field. Section 7 concludes the paper by discussing the findings and implications. Finally, Section 8 outlines potential avenues for future research.

## 2. Method Overview

Our approach to knowledge refinement is inspired by human cognitive science. When an individual fails to execute a frequently performed task, s/he will try to discover the reason and the first question that comes to mind is: "What did I do differently this time that led to failure?". Human action control is an integration of feed-forward and feedback components [37, 16]. For example, to pick up a pan one does not need to know its exact weight in advance; this can be easily determined by picking it up and slightly increasing the exerted force until the pan leaves the surface. According to Schmidt [37], human action control is hybrid, combining both feed-forward and feedback components. Schmidt argues that action schemas are set by specifying the relevant attributes of that action while leaving free parameters to be specified online while collecting environmental information. This indicates the integration of offline task planning with online sensorimotor and knowledge updating.

Improvements in the pre-specified domain knowledge are driven by the robot's failures and successes. This is achieved by connecting the robot architecture's various abstraction levels so that information about the cause of execution failure can be used to update the robot's knowledge. To do so, we developed the HDJ robotics architecture (shown in Figure 1) with task planning at its top level.

Task planning is defined as the use of a plan generation engine to solve a planning task stated as a problem instance of the form *(Os, Init, G)*. This is a triple consisting of objects *Os*, the initial state *Init* and the goal *G* specification. The problem instance is underpinned by a domain model which contains abstract parameterised structures (sometimes referred to as "operators" in the literature) representing groups of concrete actions. Hence, the word "action" is used to refer to an instantiated action structure (operator) that can be dispatched to robot control to be executed, but is defined by this parameterised action structure. Given an action's definition *P* in the domain model, each *P* consists of a pair *(pre(P), efc(P))* where *pre(P)* represents the action's preconditions, and *efc(P)* represents the effects of the actions.

Relations and properties used within the components of

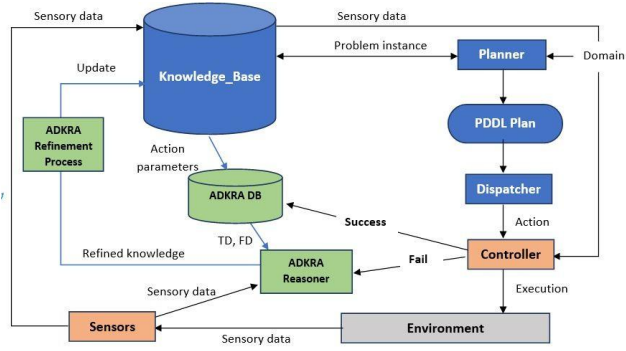

**Figure 3:** HDJ data flow diagram. The existing components (Blue boxes) and ADKRA components (Green boxes)

---

**Algorithm 1:** Anomaly detection and knowledge refinement

**Input Data:**
TD: list of m vectors of size n
FD: vector of size n
Status: KB refinement status
**Output:** Refined knowledge;
// Anomalies: List of detected anomalies
// Out: A detected outlier
// LV: Learned value
// SD: Successful action data

1 **while** *planExecution* **do**
2    **if** *not (actionExecutionFeedback-success)* **then**
3       *Anomalies ← Anomaly-detection(FD, TD)*
4       // Implementation of Algorithm 2
5       **if** *Anomalies-count > 0* **then**
6          // If Outliers exists
7          *Out ← Refinement-rule(Anomalies)*
8          // Implement refinement rules
9          *LV ← Learning-process (Out)*
10         // Learn new success values
11         *RK ← Refinement-process (LV, Out)*;
12         // Refine the KB, Algorithm 4
13         // resulting algorithm output
14    **end**
15    **else**
16       *addTrainingData (SD)*;
      // Add successful action data to the Training data
17       **if** *LV == successValue* **then**
18          *Status := Confirmed*
         // Update the KB (confirmed update)
         // refer to Figure 4
19    **end**
20 **end**
21 **end**

---

actions are contained within the robot's knowledge of the environment, encoded within a knowledge base (henceforth referred to as KB). The KB contains facts (encoded in the PDDL language [31]) representing the environment's state as well as the domain model.

HDJ has components that reason with experience-based data, leading to the refinement of the KB held in a model, as shown in Figure 2. The HDJ software includes the anomaly detection and knowledge refinement subsystem (ADKRA), which uses data from plan execution to drive the formation of refinements used to update the KB. The robot's operational experience is stored in datastores, which constitutes ADKRA's training data (TD). TD is the history of successful executions, typically maps of action model parameters, and contains all the relevant information required in the implementation of ADKRA.

We use the term "anomaly detection" here in the context of an unexpected event which is the failure of the physical execution of the plan. In this case it means identifying data that deviates from the norm and raises concerns about a particular issue or problem which could be the cause of the failure [46]. It has been utilised in various fields such as detecting bank fraud, identifying medical problems, monitoring systems, detecting ecological disruptions, ensuring network security, and recognising events [19, 39].

Anomaly detection is proposed as a novel reasoning method that draws inspiration from cognitive science, specifically the integration of feed-forward and feedback components. This approach guides the knowledge refinement process by emulating how humans address failures in frequently performed tasks.

## 3. The HDJ hierarchical system

HDJ consists of three layers (shown in Figure 1) embedded in an environment that enables the recording of planning and acting experiences, and subsequent adaptation of knowledge. The architecture we employ is inspired by previous work on robotics architectures such as the layering used in the functional architecture within a UAV [12]. It contains original components as well as existing components—in particular parts of ROSPlan [7] are utilised in the top level of HDJ. The intermediate level is a reactive level that processes actions to be executed and feedback from execution. It also includes the action dispatcher, action interface and extension components (the green boxes in Figures 1 and 3). For more details of the architecture, and its implementation in the NAO robot, the reader is referred to [17].

To illustrate the role of knowledge refinement, consider an example derived from our experiences working with a robot that performs tasks in a Railway Rolling Stock maintenance lab [30]: an *unscrew a bolt* task. Assume that an action's parameters refer to a vector such as $X = x_1,...,x_n$, where *n* is the number of attributes of the action. The action 'unscrew-bolt' has the following parameters: bolt $X_1$, unscrew-tool $X_2$, and robot $X_3$. The precondition specifies the conditions that must be met for the action to be applicable: *the robot must be holding the tool, the robot must be at the same location as the bolt, the unscrew-tool must be compatible with the bolt head* and *the bolt must be screwed in*. During the execution of an action derived from this action, a failure occurs. From the observations derived from sensor feedback, the reasoning function identifies the presence of grease as an anomaly that was not encountered in previous successful executions. As a result of the slippery bolt head, the action of unscrewing the bolt fails to achieve its intended outcome, and the bolt remains in its position. This anomaly prompts the possibility of inferring the extra precondition of *clean bolt head* within the unscrew-bolt action.

It is important to note that the focus of past work in using learning in automated planning has been on *propositional* representations, that is on learning the truth of propositional conditions [22]. While that is useful is some cases, here we concentrate on the more general approach of learning in the context of objects with numeric attributes, which is more appropriate for robot operations. In the example above where the proposition *clean bolt head* would be proposed to correct the unscrew action, in our system the "clean" attribute would be given a numeric representation rather than a propositional one, and what would be proposed to fix the action would be some measure of cleanliness (rather than true/false).

This refinement activity in HDJ starts after execution has failed and the feedback of the failed execution has been sent to the KB, as shown in Figure 2. The algorithms that perform the anomaly detection and knowledge refinement steps within HDJ are described in the sections below.

## 3.1. The Algorithms in HDJ
*Algorithm 1*

Algorithm 1 is the top-level algorithm which collects data and drives the changes to HDJ's knowledge (see Figure 3) by processing the feedback coming from the control system after each execution. The control system sends feedback to the KB and this feedback is either that the execution was successful or failed. In the latter case (Line 2), information about the failed execution will be tested to extract the differences (Line 3) by employing Algorithm 2.

The output of Algorithm 2 is a record of the extracted differences that are considered as the potential cause of failure (Line 5). This report serves as a basis for learning success values (Line 9) and subsequently refining the KB through the execution of Algorithm 4, responsible for updating the KB (Line 11). However, the process of knowledge refinement is aided by refinement rules (Line 7), which are evaluated prior to the learning process. These rules play a role in determining which attributes should be learned and refined when a group of anomalies is detected. An "Outlier" is selected from the generated Anomalies, and used in the Learning and Refinement processes.

Initially, updates are treated as temporary modifications and are further outlined in Algorithm 4. In the case of a "success" feedback (Line 16), the current success values are added to the training data, categorised as "Normal data". If the success values align with the learned value (Line 17), the status of the corresponding knowledge is set as "Confirmed", see Figure 4.

**Training Data**

The training-data (TD) can be defined as a record of all previous successful executions. Let $A = a_1, ..., a_n$ be the set of attributes that represents all information related to an action. Hence TD is a list of m vectors of size $n$, where each vector is an ordered list of attribute-values. Alternatively, TD can be represented as an $m \times n$ matrix where the columns denote $n$ attributes, and the rows maintain the values of these attributes over $m$ executions. As can be seen from the Figure 5, if we recorded 100 successful executions, then $m$ will take any value between 1 and 100.

*Algorithm 2*

Algorithm 2 outputs the patterns in data that differ and raise suspicions about a specific problem or issue [46]. The purpose is to discover the suspicious values in the failed execution, focusing on the outlier values which are often responsible for that failure.

A standard approach for AD in datasets is to create a normal data model and compare/test records against normality [41, 46]. First, we define normality for the given data. Because the TD represents the values of successful executions, we assume that historical records of successful execution contain data values falling within a normal distribution. The failed execution information (FD) is a record of a failed complete action execution that is given during operation. Each instance of FD is an ordered list of attribute-values, represented as a vector $X = x_1,...,x_n$, sharing the same form as vectors of TD. The online AD problem is to decide for each given vector $X$, whether or not $X$ is anomalous with respect to TD. Our approach, implemented in Algorithm 2, is to find anomalies - the attributes of FD that have values which do not appear in TD for the same attribute.

| Algorithm 2: Specification of Anomalies |
|---|
| **Input Data:** |
| n : integer |
| m : integer |
| TD: list of m vectors of size n |
| FD: vector of size n |
| **Output:** |
| Anomalies: Set of (index, attribute, value) |
| 1   Anomalies |
| 2     ← { (i, v(i).attribute, v(i).value): 1 ≤ i ≤ n & |
| 3         ∄ j, 1 ≤ j ≤ m, : TD(j,i).value = v(i).value } |

An outlier is a member of the set of Anomalies which is input to the Learning and Refineent Processes. Outliers can present themselves in different forms, and the anomaly detection survey [9, 1] highlights several anomaly detection methods based on different categories mainly point anomalies, contextual anomalies, and collective anomalies [1, 19].

Hodge and Austin [19] conducted a survey of outlier detection and identified three fundamental approaches. The Type 1 approach involves detecting outliers without prior knowledge of the data, similar to *unsupervised* clustering. This approach assumes that anomalies are distinct from normal data and considers them as potential outliers. On

Short Title of the Article

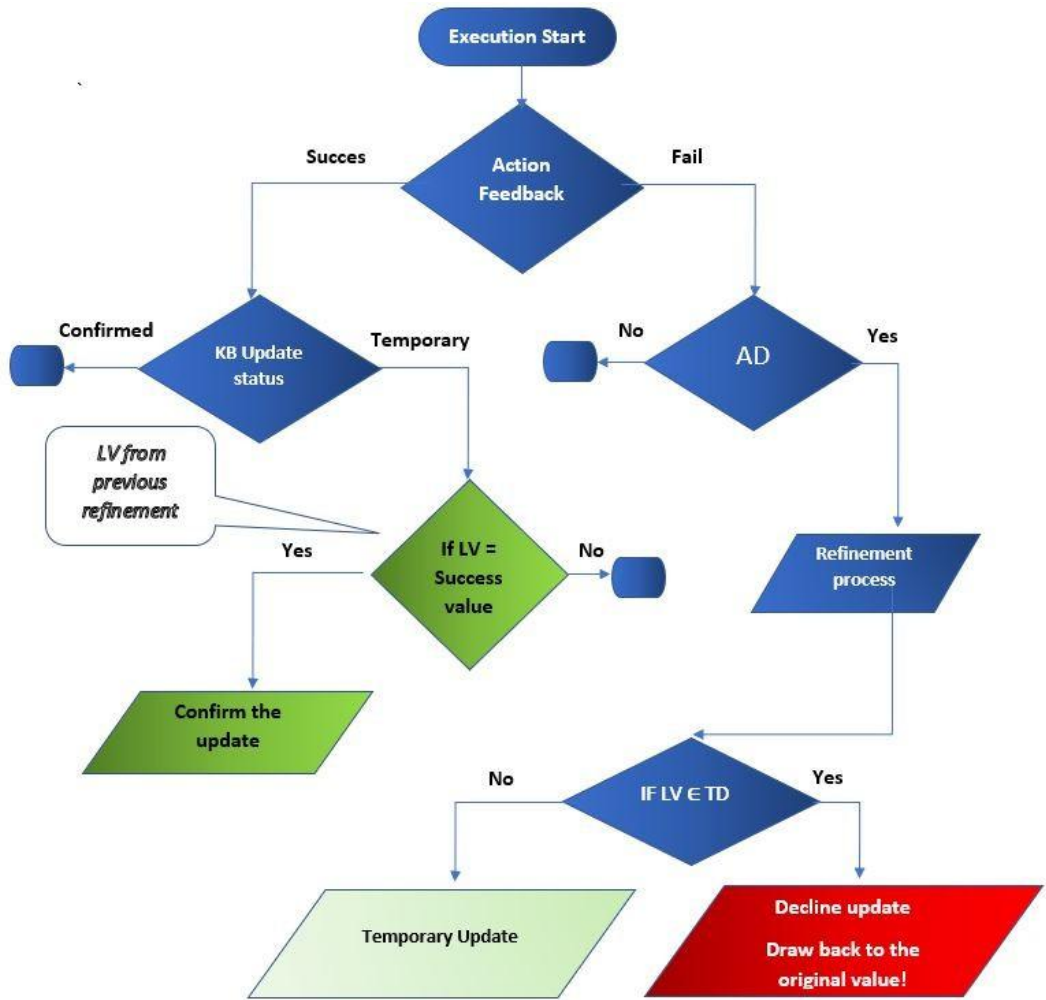

**Figure 4:** Flowchart for refinement validation: Implementation of Algorithms 1 and 4.

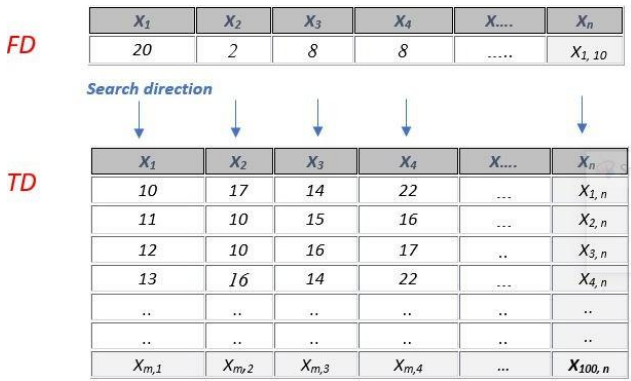

**Figure 5:** Algorithm 2 search. Each attribute instance ($x$) of *FD* is tested against its $X$ vector in TD. $n$ is number of attributes and $m$ is number of records in TD

the other hand, the Type 2 approach models both normal and abnormal data and requires pre-labelled data. It operates in a manner analogous to *supervised* classification, where outliers are explicitly labelled as abnormal. The Type 3 approach is considered a *semi-supervised* approach, where only the normal class is taught, and the algorithm learns to recognise abnormalities based on deviations from the learned normal behaviour.

The approach we adopted is semi-supervised, using initial action-based models, which resemble a supervised approach as they provide predefined information. However, there is also a component of unsupervised learning involved, as there may be undefined or unexplored information that could potentially contain anomalies. By incorporating both supervised and unsupervised aspects, our approach aims to capture a comprehensive understanding of the system, including both known and unknown action attributes.

In the field of anomaly detection, various techniques exist, including density-based methods, machine learning approaches, and neural network algorithms. Moreover, multiple anomaly detection techniques can be applied simultaneously to the same dataset. A recent trend in anomaly detection systems is the use of hybrid techniques that combine multiple methods. This approach aims to leverage the strengths of each technique while mitigating their individual weaknesses and limitations, thereby maximising the overall effectiveness of the anomaly detection process [19, 8].

**Algorithm 3:** Learning process
**Input Data:**
Outlier: (index, attribute, value)
NNhg: (index, attribute, value) % Nearest neighbour
$\eta$: Learning rate
**Output:**
LV: (index, attribute, value)

1 **if** *Outlier.value > NNhg.value* **then**
2     LV ← (Outlier.index, Outlier.attribute, Outlier.value - $\eta$)
3 **else**
4     **if** *Outlier.value < NNhg.value* **then**
5        LV ← (Outlier.index, Outlier.attribute, Outlier.value + $\eta$)
         // If anomaly detection less than its nearest neighbour
6     **end**
7 **end**

if Outlier.value > NNhg.value then
    LV = Outlier.value – $\eta$
    [Example 1: LV= 24 -1 ← 24 > 20]
else
if Outlier.value < NNhg.value then
    LV = Outlier.value + $\eta$
    [Example 2: LV= 13 +1 ← 13 < 15]

**Figure 6:** Example comparisons in the learning process.

**The Refinement Process and Generalisation.**

*Algorithm 3*

Algorithm 3 derives a new success value (*LV*) based on the outlier chosen from the anomalies output from Algorithm 2. The goal is to minimise the discrepancy between the outlier's value and its closest neighbour's value in the training data. *LV* is calculated by adding a learning rate, $\eta$, to the value of the detected outlier's value (Lines 2 and 5). The learning rate represents a step towards the nearest neighbour, *NNg*, and serves to reduce the discrepancy between the predicted value (initial value in the KB) and the desired values (the observed experiences). This is illustrated in Figure 6.

In machine learning and statistics, the step size or learning rate parameter in an optimisation algorithm determines the magnitude of the change at each iteration as it descends the cost landscape in the direction with the steepest descent [34]. As seen in Figure 6, the direction in which the step is taken is determined by the comparison of *Outlier.value* and *NNhg.value*. The learning process, *LP*, considers two comparisons of outliers. If the value of *Outlier* is greater than the value of *NNg* in TD, the new value will be (*Outlier.value* minus $\eta$), otherwise the new value will be (*Outlier.value* plus $\eta$). Then *LV* will be passed to *Refinement-process* to refine the related value in KB.

*Algorithm 4*

Algorithm 4 implements the knowledge refinement process. The refinement process is in charge of updating the

**Algorithm 4:** Refinement Process
**Input Data:**
LV: (index, attribute, value)
Outlier: (index, attribute, value)
P: Action
Status: KB refinement status
TD: list of vectors
**Output:** Refined action knowledge

1 **if** $LV \in TD$ **then**
2     exit // Refinement rejected
3 **else**
4     *Update KB (LV.index, LV.attribute, LV.value, P);*
       // Refine the value of action attribute associated with the detected outlier)
5     *Status: = temporary*
       // Update the KB (temporary update)
       // Refer to Figure 4
6 **end**

KB, which holds instances of values in the PDDL language. After *LV* is learned, by implementing *LP* (Algorithm 3), it is passed to Algorithm 4 to update the relevant attribute in KB. Any detected anomaly is a value that indicates parameters or pieces of information in the KB, such as pre-conditions and states.

The updates are initially considered temporary and are later confirmed as permanent once they have been validated. The refinement process continues in iterations until successful execution is achieved, at which point the update is confirmed, as Figure 4 illustrates. The refinement process incrementally updates the KB based on experience and each time the model reduces the gap between the outlier and its nearest neighbour in the TD, the new solution is validated before confirming the update as valid (Line 1).

The validation process checks and compares *LV* against *TD*. In the case of the current failure, if $LV \in TD$, the update will be rejected, and the system will revert to the last confirmed update (Line 2). When $LV \in TD$, it signifies that the newly learned value, intended to prevent failure, is already recognised as a successful value. However, given the consistent failure of the system to execute the action, this suggests that the detected anomaly may not be the primary cause of the action failure or there might be another underlying reason. Consequently, it is advisable for the system to reject this *LV* and revert back to its original values.

*Refinement Rules*

The refinement rules are implemented in line 7 of Algorithm 1. The number and selection of KB instances to refine in each iteration is domain-specific. While *Algorithm 2* has the capability to detect multiple anomalies, it is important to consider the relationships between the attributes of the domain model when refining its knowledge due to the mutual dependencies observed among the action attributes. For example, if attributes $X_1$ and $X_2$ exhibit an inverse relationship ($X_1 \propto X_2$), where the values of $X_2$ follow the pattern of $X_1$, then the $X_2$ attribute is selected for refinement. This categorisation is referred to as a 'collective' anomaly.

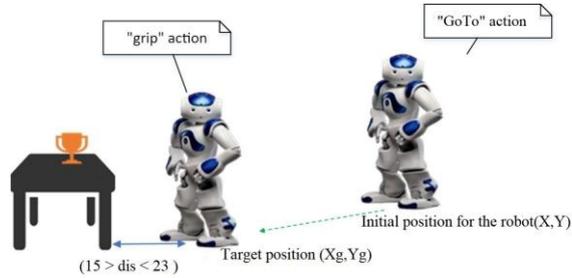

**Figure 7:** Kitchen scenario - gripping Task.

*Method Summary*

We summarise our approach by stating the necessary steps which would allow the method to be generalised to other application areas.

An engineer has to create an initial domain model of a robot's environment and its own capabilities via a domain model, and supplement this with a database structure for capturing experience data, and with learning parameters for each attribute in the model. Successful action execution data must be directed into this database, building up robot experience based on real executions. In the event of a failed execution, an anomaly detection algorithm has to be applied to identify anomalies within the failed execution data. It is advisable to employ a combination of multiple methods to detect anomalies pertaining to the same attribute. This strategy helps overcome limitations inherent to individual methods and allows for the exploitation of their respective strengths.

Next, new attribute-values have to be derived based on the detected anomalies, which are likely to lead to generated plans with successful execution. The objective in choosing values is to minimise the gap between the outlier and its nearest neighbour in the data. Finally, the KB is updated based on these learned values.

## 4. Case Study

As the primary method of evaluation, we will use a kitchen scenario [32] as a case study. As shown in Figure 7, the tasks performed by the robot included gripping a cup and moving it from one table to another. As Figure 8 represents, a successful execution of the 'grip' action is when the robot can carry out its actions; in this specific case, this means the robot can reach for and grip the cup.

The grasping task requires several functions, such as vision recognition, object detection, distance calculation and motion design. Distance measurement is of great importance in this example as the accuracy of the distance data affects the performance of the robot and lead to execution failure. In Figure 7, we can see that for a successful execution, the distance between the NAO robot and the table must be from 15 to 23 centimetres. If this range is specified incorrectly, e.g., 15 to 27 centimetres, it will result in some failed executions. Hence, if the robot is equipped with an autonomous reasoning function that enables it to realise this fault in the KB and justify and correct the model's description, this will make for more robust and longer-term autonomy.

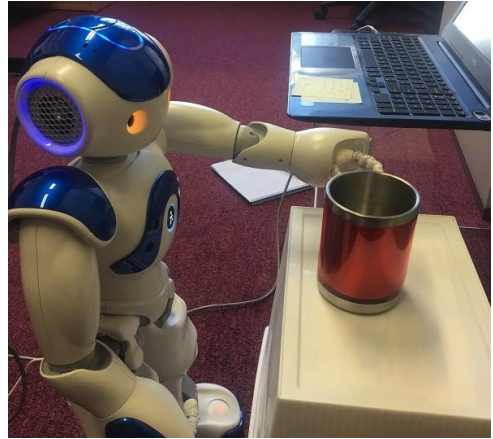

**Figure 8:** Successful execution: the NAO robot can reach for the cup.

In this research, we use the NAO robot in our experiments. The NAO robot can learn, recognise and track objects, as well as perform various motion tasks. To perform the grasping task, the planning system dispatches two actions ('goto' and 'grip') to the controller of the robot. Domain model definitions and action parameters, residing in the KB, are provided in the following subsections.

**Parameters and Attributes**

The action 'grip' involves several attributes, and in this particular scenario, we have identified three key attributes. The distance between the robot and the table represents its first attribute $X_1$, while $X_2$ is the angle attribute that represents the position of the targeted object. Furthermore, the relationship between the 'distance' and the 'angle' attributes is another factor, a correlated attribute [35], must be considered as an attribute of the grip action that can be represented as $X_3$. These action parameters can be learned and justified based on online execution.

### 4.1. The PDDL Model

To be able to do task planning in our architecture, knowledge must be encoded in the form of a domain model, and for this study, a simple (non-temporal) variant of the PDDL description language v. 2.1 [14] was used. The PDDL model of the gripping task is represented in Figure 9.

In the Domain Model, the functions *dist-to* and *hwangle* are associated with the preconditions of the action *grip*. These 'fluents' (variables in PDDL) are employed to represent the distance rule (*dis*) and the angle rule (*HeadYawAngle*). Incorrectly specified preconditions will cause the action to fail, which, in turn, will affect the constraints on the preceding 'goto' action. Doing so could lead to failure to

```
(define (domain NAO)
(:requirements :strips :typing :fluents)
(:types waypoint thing robot gripper)
(:predicates
      (atrobby ?r - robot ?x - waypoint)
      (pos ?o - thing ?x - waypoint)
      (free ?r - robot ?g - gripper)
      (carry ?r - robot ?o - thing ?g
      - gripper))
(:functions
      (dist_to ?x1 - waypoint ?x2 - waypoint)
      (maxdis ?grp - gripper)
      (mindis ?grp - gripper)
      (hwangle  ?r - robot)
      (maxhwangle ?r - robot)
      (minhwangle ?r - robot))
(:action goto
 :parameters (?r - robot ?from - waypoint ?to
 - waypoint)
 :precondition (and
            (atrobby ?r ?from))
 :effect (and
            ( a t r o b b y   ? r   ? to )
            (not(atrobby ?r ?from))))
(:action grip
 :parameters (?r - robot ?obj - thing
?waypoint1 - waypoint
?waypoint2 - waypoint ?g - gripper  )

 :precondition (and
      (atrobby ?r ?waypoint1)
      (pos ?obj ?waypoint2)
      (free ?r ?g)
      (>(dist_to ?waypoint1 ?waypoint2)
         (mindis ?g))
      (<(dist_to ?waypoint1 ?waypoint2)
         (maxdis ?g))
      (<(hwangle ?r)(maxhwangle ?r))
      (>(hwangle ?r)(minhwangle ?r)))
  :effect (and
      (carry ?r ?obj ?g)
      (not (free ?r ?g)))))
```

**Figure 9:** A segment of the KB's PDDL model implemented in our experiments

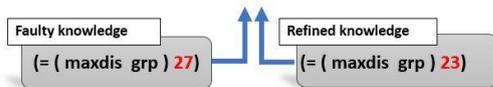

**Figure 10:** Knowledge refinement of the distance constraint

execute a seemingly valid abstract plan. Figure 10 demonstrates the refinement of the distance rule (*dis*) as the fluent (*maxdis*) is corrected from 27 to 23.

```
; Cost: 0.001
; Time 0.00
0.000: (goto nao wp0 wp2)    [0.001]
0.001: (grip nao redcup wp2 wp1 grp)   [0.001]
```

**Figure 11:** The plan generated from faulty domain knowledge resulting in execution failure.

```
; Cost: 0.001
; Time 0.00
0.000: (goto nao wp0 wp4) [0.001]
0.001: (grip nao redcup wp4 wp1 grp) [0.001]
```

**Figure 12:** The plan generated from refined, more accurate encoding of domain knowledge resulting in successful execution.

Our example illustrates the possibility that plan rules and pre-conditions could be updated to avoid failure of future executions. This can be done by repairing incorrectly specified actions and recognising new attributes or relationships [25].

The planner's responsiveness to knowledge refinement is clearly demonstrated through the improved plan representation and successful execution outcome depicted in Figures 11 and 12. Figure 11 presents the plan generated from a faulty domain, which resulted in a failed execution while Figure 12 shows the plan generated after the domain knowledge has been refined, leading to successful execution. The difference between the two illustrates the positive impact the revised knowledge has on the planner's performance.

## 5. Evaluation

A set of experiments was designed to test the effectiveness of HDJ in dealing with a flawed model of its environmental capabilities. We explore its behaviour within the Case Study, explained in Section 4. In the experiments, the model contained faulty domain knowledge, resulting in some of the generated plans failing on execution. We tested the effectiveness of HDJ software in reducing the rate of failed executions as a result of knowledge refinement, and can claim a baseline success when this measure reduces monotonically over time.

Each experiment investigates one or more action parameters that were incorrectly specified, or ignored, in the KB. Typically, such incorrect parameters and attributes tend to form types of anomalies:

- Distance failures: the distance between the robot and the table.
- Angle failures: the angle of the position of the cup on the table to the robot position.
- Combinations of attributes failures: where specific combinations of distance and angle attributes cause failure.
- Group of attribute failures: where more than one attribute caused the failure. e,g.,the distance and angle attributes.

## 5.1. The Testing Procedure

When an action within a plan that has been generated to solve a given task fails to execute, the robot must identify the cause and then update to the knowledge base. The ADKRA subsystem implements the anomaly detection and refinement processes reasons about the cause of the failure and then refines the KB based on the outcome, to repair the cause of the failure. To test this and evaluate ADKRA, and hence HDJ's effectiveness in reducing failure rates, we used the following procedure in each of the four kinds of tests.

1. Apply a change to the KB to make it faulty (e.g., give it the wrong maxdis).
2. Generate 100 problem files where the parameter/s in the KB are *randomly* generated to give a different problem each time (e.g., different distances between way points, or/and different initial states of the robot and cup)
3. Generate a plan for each of the 100 problems. To do this we use a general PDDL+ planning engine ENHSP [13], which has been embedded in HDJ.
4. Execute each of the 100 plans:
   - If the execution succeeds, add the execution data to TD.
   - If the execution fails, and ADKRA is not being used, then record the failure.
   - If the plan fails, and we are intending to use ADKRA, then record the failure; run ADKRA to do anomaly detection and knowledge refinement on the KB. Store the newly refined knowledge for use in later tests.
5. If we are using ADKRA, then after the 100 tests, REPEAT STEP 3, and execute each of the 100 plans generated, recording the total number of failures.

Executing this procedure for each of the four aspects above, we can determine how much using the ADKRA process helps reduce failure rates within the execution of generated planning solutions. This is done by comparing the number of failures without ADKRA, with the number of failures in step 5 above, when ADKRA was used.

## 5.2. Development and Technical Setup

The development setup for the experiments encompasses two key developmental aspects: the integration of new and existing components within HDJ and the design of the current state facts and domain model. First we describe the relevant aspects of the PDDL language employed in our experiments then outline the software development process and the integration of middleware components.

## 5.3. Planning Domain Definition Language

To capture both the symbolic and numerical aspects of robot domains, we chose to employ PDDL version 2.1 which allows numeric fluents (shown in Figure 9). It is inspired by the domain model of 'pouring water between jugs", in PDDLv2.1 [14].

Numeric fluents are used in the preconditions of the 'grip' action, and the metrics are described in the problem file. The numeric expressions are values associated with tuples of domain objects by domain functions and they allow the setting of different values that construct different solutions to problems of the same domain [14].

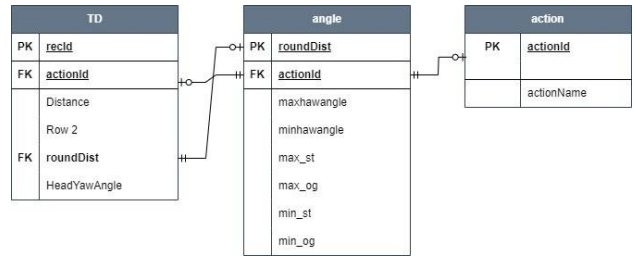

**Figure 13:** A part of the ADKRA database design, which includes a representation of the relationships between action attributes.

Given action *P* from the domain model, *PNE* is a primitive numeric expression (*fluent*) of a task related to the precondition of *P*. The execution of Algorithm 1 detects the cause of a failure, learns success values *LV* and then uses the learned value to update the KB, by replacing the original value of expression *PNE* with the learned value *LV*.

The PDDL2.1 models contain objects, and actions are represented as parameterised operations which change the status (attribute-values, relations) of objects. Numbers are used as values of attributes of objects and are manipulated through their connections with the objects that are identified in the initial state [14].

In the model of the Case Study scenario, the functions *dist-to*, *mindis*, and *maxdis*, are referred to as numeric fluents, and are associated with the preconditions of the action *grip*. These fluents are employed to represent the distance attribute and its permitted range of values. We use a prefix syntax (comparison) for the *grip* action and its precondition (>=(dist-to ?r ?waypoint) (mindis ?r ?waypoint)). This forms a plan rule we call *dis* which is similar to the distance rule. We also have a *HeadYawAngle* rule where we employ numeric fluents in the functions *hwangle* and *maxhwangle*.

## 5.4. Software and Middleware

HDJ system development entails implementing a new programme integrated with the existing components of ROS. In this research, we utilise ROS and ROSPLan as components of HDJ architecture.

The controller of NAO was developed as a ROS node; this is because the controller communicates with ROSPlan to execute the planned actions. The ENHSP planner [13] was employed in our experiments to generate plans.

The reasoning and refinement algorithms proposed by this research (ADKRA) have been developed and embedded in the context of NAO robot and HDJ architecture.

The HDJ system utilises the SQLite database to store and retrieve knowledge that is used to construct the initial state of the problem instance. This database is created to include structured information of the KB such as the actions parameters and all related information including the relationships between the attributes and correlated attributes. This includes the actions' pre-conditions, plan rules and even the representations of the targeted object. This provides the flexibility to customise the fluents values before passing them to the system.

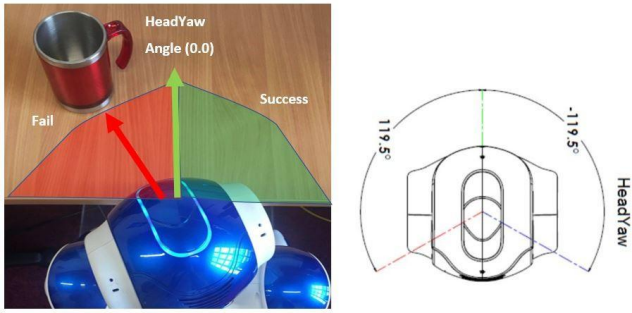

**Figure 14:** Success and fail ranges of 'HeadYawangle', to grip using the right hand of the NAO robot.

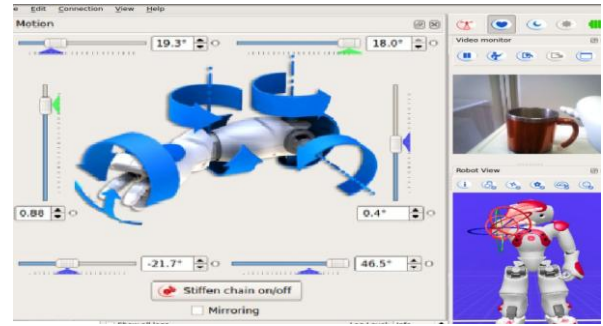

**Figure 15:** The joint values of NAO's arm in grasping position at the maximum degree towards the body.

Figure 13 shows the representation of the relationship between the *HeadYawAngle* rule and the *dis* rule as in the database structure. The sensory data holds values of action attributes such as the current distance to the object and its location. Furthermore, the information of the targeted object, the red cup, is also processed during the instanstiation process, [23], as it is used to detect the object and calculate the distance and the angle attributes. The Python programming language was used for controller development.

In the domain model, represented in Figure 9, we employ numeric fluents in the new pre-conditions of the 'grip' action. All values of fluents are saved in an external file by the HDJ system and are passed to the problem file through the instantiation process and to the system through the execution stage. The initial values of *mindis* and *maxdis* are calculated and processed by the HDJ system and are stored as an external file to be read and retrieved by the system. Meanwhile, the values of *dist-to*, and *hwangle* are sensory data (actual distance and angle values) that are measured online during execution stages.

### 5.5. Experimental Setup

This section describes the experimental setup for four types of experiments. Each setup is described based on investigations conducted on each failure type, and the implementation setup for each experiment.

*Investigating Distance Failures*

These experiments investigate the distance parameter as a cause of failure, particularly the distance between the robot and the cup as seen in Figure 7. For the successful execution of gripping, the distance must be in the range of 15 to 23 centimetres (see Figure 7). If this range is extended (e.g., 27cm>dis>15cm), some failures would be caused during execution because these extended values (24, 25, 26 and 27) of the distance will cause the execution to fail.

In order that some failures would be caused during execution, faulty knowledge was given to the system in this set of experiments. This consisted of an erroneous distance precondition as follows: the correct range of (23cm>dis>15cm) was extended to (27cm>dis>15cm). For each experimental episode the initial position of the robot is changed, and then the generated solution plan for this new gripping task is checked for success or failure in execution. If the task plan fails, the KB is refined using output of ADKRA (which is produced after a number of refinements) of these experiments.

*Investigating Angle Failures*

These experiments investigate the angle parameter as a cause of failure. For a successful execution of the gripping task, by the right hand, this angle should be in a range of 0.0° to -25° (see Figure 14). If this range is extended to a wider one (e.g., 25°>HeadYawAngle>-25°), some failures would be caused during execution because these extended values (greater than 0°) will cause the execution to fail.

The NAO robot has 25 degrees of freedom (DOF), of which 14 are for the upper part including its trunk, arms, and head, including the rotation axis of its arms, and head joints which are inclined at 45° towards the body. The head can rotate about yaw and pitch axes. Each arm has 2 DOF at the shoulder, 2 DOF at the elbow, 1 DOF at the wrist, and 1 additional DOF for the hand's grasping.

The available DOF of the arms of the NAO robot limits its abilities in grasping. In particular, the angle of the object's position is a critical condition for successful grasping. As seen in Figure 15, the maximum rotation of the shoulder joint towards the body is only able to bring the arm in front of the robot's cameras. The ideal position for grasping an object is when the HeadYawAngle= 0.0.

In order to cause some failures during execution, faulty knowledge was added, containing an erroneous HeadYawAngle precondition as follows: The correct range of ( 0.0° > HeadYawAngle > -25°) was extended to (0.0° > HeadYawAngle > -29°). For each experimental episode, the initial position of the robot was not changed, but the angle of the cup position on the table was randomly changed (see Figure 14) and the distance was fixed to 18 cm; Then the generated solution plan for this new gripping task was checked for success or failure in execution. If the task plan failed, the KB was refined using ADKRA.

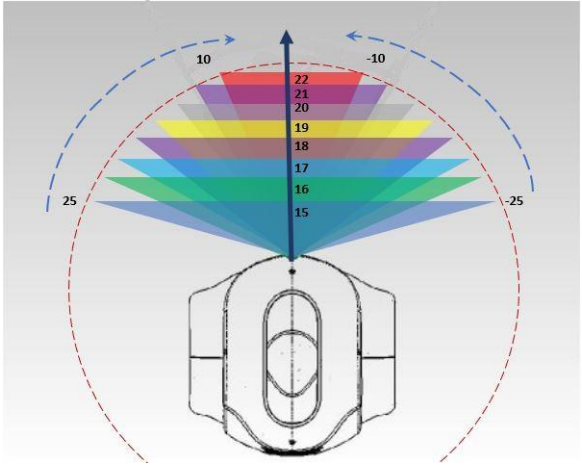

**Figure 16:** The relationship of the distance and angle attributes.

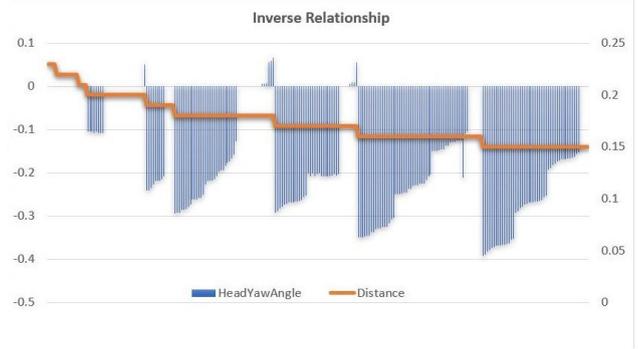

**Figure 17:** TD distribution. Distance and HeadYawAngle data

*Investigating Attribute Combinations Failures*

These experiments investigate the particular combinations of action attributes (Distance and HeadYawAngle) that cause executions to fail. For successful grasping, our experiments have shown that each distance point accepts a specific range of angle. The error in the KB is to ignore the associated attribute, which is the inverse relationship between the distance and angle parameters of the 'grip' action. As seen in Figure 16, when the *Distance* increases, the *HeadYawAngle* range is narrowed and goes towards a *0.0* angle. Figure 17 represents the inverse relationship between the *HeadYawAngle* attribute and the *Distance* attribute. For example, the 15 centimetre distance accepts the full range (0.0>HeadYawAngle>-25), for the right hand. However, the 20 centimetre distance accepts a shorter range of angle of about (0.0>HeadYawAngle>-12).

For this reason, particular combinations (of the *Distance* and the *HeadYawAngle* attributes) cause execution failure. For example, despite the fact that the ranges (0.0 >HeadYawAngle > -25) and (23cm > dis > 15cm) are considered successful ranges of distance and angle attributes, some values within these ranges, if combined, will cause execution to fail. A 'Collective' anomaly is a group of attribute instances that appear isolated compared to the rest of the data-set. A specific case of 'Collective' is when each single attribute is not an anomaly but as a group appears together as an anomaly [9]. These experiments test such cases.

In order to cause some failures during execution, faulty knowledge was given to the system in this set of experiments. In the domain model part of the KB, represented in Figure 9, we have two preconditions that accept a range of values of the *Distance* and *HeadYawAngle* attributes. We set the ranges of both rules *HeadYawAngle* and *dis* to (0.0>HeadYawAngle>-25) and (23cm>dis>15cm); These ranges are considered to be the range needed for success.

However, we know that within this range specific cross values will fail the execution; this is due to ignoring the inverse relationship between the *Distance* and *HeadYawAngle* attributes, which is considered as a fault in the KB. For each episode, we changed both the initial position of the robot to the table, and the angle of the cup position on the table as well.

The correlation between the angle and the distance attributes is considered an attribute of the action 'goto' that needs to be checked as well when execution failed. To detect such an anomaly, we may have to create a new attribute (a correlated attribute [35]). Employing the 'Collective' AD [1, 19] can help detect unusual combinations of values that would not have possibly been detected by earlier steps of Algorithm 1.

For each failed execution, the ADKRA processes (Algorithm 1) learns values *LV* and used them to refine the module. We stated in Section 6 that the relationships of the domain's attributes determine which KB attribute is to be refined. So, for this particular instance, the KB attribute (*maxhwangle*) to refine is based on its relationship with the *Distance* attribute. The angle is classified as a 'slave' relationship, as seen in Figure 13, with the distance attribute. The refinement system is set to only learn and refine the *HeadYawAngle* attribute.

*Investigating Failure of a Group of Attributes*

These experiments investigate execution failures due to more than one attribute (e.g, angle and distance) being specified incorrectly. For example, a distance of 24cm is considered a failure value and -27° is an angle failure value. A 'Collective' anomaly is where a group of attributes— instances appears isolated compared to the rest of the data-set [9]. This experiment is the general case of 'Collective' anomalies, when detecting two or more attributes as being anomalies.

In order to cause some failures during execution, the faulty knowledge given to the system in this set of experiments contained erroneous distance and HeadYawAngle preconditions as follows: The correct range of (23cm > dis > 15cm) was extended to (25cm > dis > 15cm) and the correct range of (0.0 > Head YawAngle > -25) was extended to (0.0 > HeadYawAngle > -27). These ranges are both incorrect and lead to failure during the execution. For each experimental episode, the distance of the robot to the cup was randomly changed, while the angle of the cup position was fixed at -20°. Then the generated solution plan for this new gripping task was checked for success or failure in execution. If the task plan failed, the knowledge was refined using ADKRA.

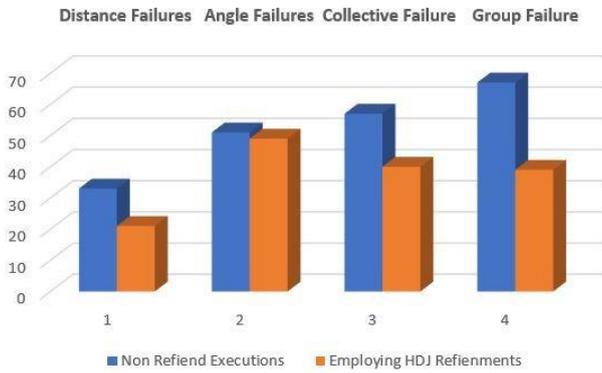

**Figure 18:** Field executions with and without knowledge refinement by ADKRA

### 5.6. Experimental Process and Results

The test results demonstrate the outcomes obtained from executing the specified testing procedure. As the ADKRA subsystem adjusts the parameters, the failure rate gradually decreases over time. The comprehensive outcomes of these tests are depicted in Figure 18. Evaluation of the predictions generated by the ADKRA reasoner involves the utilisation of key evaluation metrics, including false negative rate FNR, true positive rate TPR, Accuracy, and Precision, as outlined in Table 1. These metrics serve as vital benchmarks for assessing the model's ability to produce accurate predictions. By providing insights into different aspects of the model's performance, these metrics enable the evaluation of its effectiveness and guide further improvements.

The experiments demonstrate the efficiency of our approach in refining the KB of the task planner. ADKRA reasons and detects anomalies efficiently and can refine the knowledge for successful future executions, reducing plan execution failures after repeated use. For example, by the end of the distance experiment, the ADKRA successfully corrected the values of the distance rule. The instance value of the maximum allowed distance (maxdis ?r ?waypoint) was corrected to be 23 centimetres.

The experiments also demonstrate that accurate knowledge can overcome issues related to inaccurate domain models. In this case, the model represented in Figure 9, ig- nores the rational relationship between the 'Distance' and 'HeadYawangle' attributes. Each distance point should have accepted a specific range of angle values, but the absence of this representation caused failing action executions due to the missing representation of parameter relationships. However, despite this flaw, ADKRA, particularly in the group attribute failures, successfully refined the range of the 'HeadYawangle' rule and set the correct range for each distance value of the 'dis' rule. This success was attributed to the ADKRA database considering the relationship between these attributes.

**Table 1**
The Overall Accuracy of Predictions by ADKRA Reasoner

| Attribute | Obs. | TP | FN | Preci. | Accu. | FNR | TPR |
|---|---|---|---|---|---|---|---|
| Distance | 65 | 61 | 2 | 96.8% | 93.8% | 3.17% | 96.8% |
| Angle | 34 | 15 | 2 | 46.8% | 44.1% | 11.8% | 88.2% |
| Collective | 39 | 25 | 7 | 78.1% | 64.1% | 21.88% | 78.1% |
| Collective (Group) | 36 | 24 | 9 | 88.8% | 66.6% | 27.2% | 27.7% |

Notably, ADKRA serves as a domain validator, providing an additional advantage in anomaly detection. Specifically, in the case of the 'Collective' anomaly, where the combination of attributes is deemed anomalous while each individual attribute alone is not, it unveils a mutual relationship between the attributes. This type of anomaly detection can be utilised to unveil previously unrepresented relationships in domain models. Disregarding such relationships can lead to an increase in false negative (FN) predictions. Therefore, incorporating the 'Collective' anomaly detection approach aids in capturing these intricate attribute relationships and mitigating the risk of false negatives.

Furthermore, the experiments also demonstrated that real-world applications introduce complications associated with the real environment and sensor issues. The results indicated that the NAO robot's measurement of the distance to objects is not currently accurate enough, which is the primary source of FN predictions observed in our experiments. The low resolution of the NAO robot's camera [45], combined with hardware specifications issues and the distance estimation method, contribute to inaccurate distance measurements. Various techniques for depth estimation are available, each with its own outcomes and drawbacks.

The accuracy of sensory data plays a crucial role in ensuring successful predictions. One challenge arises from the overlapping edges between accumulated data from successful executions and data from failed executions, as illustrated in Figure 19. For instance, the distance value of 23 centimetres is present in both datasets. While it is considered a successful attribute value in the training data (TD), it can also lead to execution failure. The presence of this value in both datasets hinders the robot from making accurate predictions, as it fails to recognise it as an anomaly in the training data.

### 6. Related Work

While many existing robotics techniques employ anomaly detection for various purposes, such as condition monitoring [24, 20, 18], novelty detection in the environment [10, 11], learn differences in object classification [27], Robot-assisted feeding in healthcare applications [36]. fault detection [20], and modelling spatio-temporal dynamics of continuous processes in the environment through frequency spectrum analysis [26], and identifying anomalous behaviours of robots [3]. There is little evidence that AD has been used specifically to identify inaccuracies in a robot's symbolic knowledge initiated by the execution failure of its generated plans, as covered in this work.

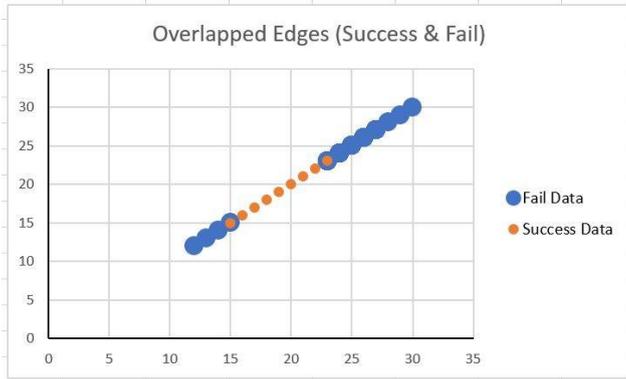

**Figure 19:** In both datasets, the distance attribute is captured, reflecting both successful and failed executions. However, there exists an overlap between the two datasets at the edges. This overlap can be attributed to inaccuracies in distance measurements stemming from the low accuracy of sensory data and the estimation method utilised.

Existing works typically refine planning domains or knowledge by identifying the cause of failure as a means to drive improvement. However, these approaches often employ different methods and are constrained to specific functions that differ from the methodology proposed in this paper. For example, LIVE [38] utilises prediction rules and the incremental enlargement heuristic to identify differences in faulty rules, while EXPO [15] refines incomplete planning knowledge using the ORM method. Similarly, the CRAM framework [5] utilises plans in the CPL language and a knowledge processing system, while GDA [44] and OB-SERVER [43] adopt learning from demonstration.

Various works have been proposed to refine planning models and address planning challenges. However, our proposed approach stands out by emphasising the utilisation of anomaly detection as a reasoning method to identify the causes of failures in plan execution and drive appropriate changes to the plans accordingly. This unique perspective adds value to the existing body of research on planning refinement in robotics. For example, Lindsay et al. [29] utilise machine learning techniques to refine hybrid domain models, while Ambite et al. [2] developed the PBR system, which leverages conflicting choice points to refine plans. Stulp et al. [40] improve the execution of planned actions by predicting their effects and performance based on real-time feedback. In addition, Verma et al. [42] created PLEXIL, an execution language that allows for richer representations of executable plans. Ziparo et al. [47] proposed the PNPs programming language which employs Petri nets to design robot and multi-robot behaviours, providing a higher-level representation that facilitates plan refinement and modification. The LearnPNP tool [28] combines the Petri Net Plans formalism with reinforcement learning, enabling learning and adaptation of plans. Furthermore, Bezrucav et al. [6] present an approach for scenarios involving teams of humans and robots, incorporating parallel planning and dispatching features to reduce re-planning waiting times.

## 7. Discussion and Conclusion

This paper has introduced the HDJ architecture, which we have implemented on a physical NAO robot. The original contribution of the paper is the description, implementation and evaluation of a method to detect causes of action execution failure due to faulty task planning knowledge, and to repair that knowledge to improve the robustness of plan generation and execution. Implementing and testing this method in a real world (rather than some abstract virtual world) is an essential factor in this contribution.

The refinement process effectively validates the system predicates and the refined knowledge before setting it as permanent. It is capable of exposing inaccuracy in the domain model, and extracting accurate knowledge from inaccurate domain models; this is due to its ability to specify and customise knowledge with the help of its database of past experience.

One limitation of this algorithm is its heavy reliance on the accuracy and reliability of the feedback obtained from the control system. If the feedback is incomplete or inaccurate, the algorithm may struggle to correctly detect anomalies, leading to incorrect learning and knowledge base updates. Moreover, this algorithm may not be suitable for tasks or systems where obtaining feedback is challenging or not feasible.

Furthermore, it is important to note that the proposed method aims to reason and predict the cause of execution failure. Therefore, the system should be capable of detecting action failures in the first place. Only after successful action failure detection can the ADKRA be executed for reasoning and knowledge refinement purposes.

The Case Study follows a supervised anomaly detection (AD) approach. This choice is based on the utilisation of an already defined action model with predefined action attributes. However, it is important to note that there may be relevant information related to the action that could potentially indicate anomalies. To discover such anomalies, the integration of unsupervised AD techniques may be necessary.

Consequently, a semi-supervised AD approach would be appropriate as it combines elements of both supervised and unsupervised AD methodologies. In this study, we utilised a simple (non-temporal) PDDL model and the efficiency of ADKRA was tested and approved with 'Point' and 'Collective' anomaly types. We also employed density based AD technique to detect anomalies in the action attribute values.

However, the low quality of NAO cameras led to overlapped data edges, which reduces the performance of the reasoning system.

## 8. Future Work

Going forward, we aim to broaden the scope of knowledge that HDJ can refine and conduct more experiments across various scenarios. This includes using a temporal-PDDL model with "Contextual" AD, which is usually employed for time series models, in addition to the "Point" and "Collective" AD types. We plan to examine other AD approaches and techniques, potentially combining multiple methods to overcome each one's limitations and take advantage of their strengths.

To enhance the performance of the ADKRA system, it is crucial to address the issue of FN that result in a high FNR due to undetected anomalies. This challenge can be attributed to several factors. Firstly, inaccurate measurements obtained from sensors introduce errors and contribute to false negatives. Secondly, the lack of representation of interdependencies among action attributes within the domain model hampers accurate anomaly detection. Additionally, undefined action attributes in the model further complicate the detection process.

One approach to address the first issue is to employ unsupervised anomaly detection techniques, such as ANN-based anomaly detection. These techniques leverage the power of artificial neural networks to improve anomaly identification and reduce the false negative rate. To mitigate the second issue, the application of 'Collective' anomaly detection proves instrumental in rectifying the inadequacy of the domain model, related to dependency between action parameters. This method captures and incorporates previously unrepresented relationships among the action attributes, enabling a more comprehensive understanding of anomalies. Finally, employing high-quality sensory systems could help reduce the impact of inaccuracies in sensory data, further improving the performance of the ADKRA system.

We are considering the adoption of a hybrid anomaly detection technique that combines vision recognition with density-based techniques. The objective is to improve the accuracy and effectiveness of anomaly detection in our scenario. To achieve this, we plan to integrate unsupervised techniques, such as utilizing Artificial Neural Networks (ANN) for anomaly detection.

For instance, the vision recognition AD component could be employed to detect anomalies or deviations in the visual characteristics of objects. This will enable us to learn and identify anomalies such as a "cleanliness" (referring to our motivating example) if this had not been pre-engineered. On the other hand, the density-based techniques would focus on detecting anomalies based on differences in the geometric information of the action being performed. By leveraging the complementary strengths of these two techniques, we anticipate improved anomaly detection capabilities, allowing us to detect anomalies from multiple perspectives and enhance the overall accuracy and robustness of our system.

## References

[1] Akoglu, L., Tong, H., and Koutra, D. (2015). Graph based anomaly detection and description: a survey. *Data mining and knowledge discovery*, 29:626–688.

[2] Ambite, J. L. and Knoblock, C. A. (1997). Planning by rewriting: Efficiently generating high-quality plans. Technical report, UNIVERSITY OF SOUTHERN CALIFORNIA MARINA DEL REY INFORMATION SCIENCES INST.

[3] Ando, S., Thanomphongphan, T., Hoshino, D., Seki, Y., and Suzuki, E. (2011). Ace: anomaly clustering ensemble for multi-perspective anomaly detection in robot behaviors. In *Proceedings of the 2011 SIAM International Conference on Data Mining*, pages 1–12. SIAM.

[4] Arkin, R. C. (1998). *Behavior-based robotics*. MIT press.

[5] Beetz, M., Mösenlechner, L., and Tenorth, M. (2010). Cram—a cognitive robot abstract machine for everyday manipulation in human environments. In *2010 IEEE/RSJ International Conference on Intelligent Robots and Systems*, pages 1012–1017. IEEE.

[6] Bezrucav, S.-O. and Corves, B. (2020). Improved ai planning for cooperating teams of humans and robots. In *Proceedings of the Planning and Robotics (PlanRob) Workshop—ICAPS*.

[7] Cashmore, M., Fox, M., Long, D., Magazzeni, D., Ridder, B., Carrera, A., Palomeras, N., Hurtos, N., and Carreras, M. (2015). Rosplan: Planning in the robot operating system. In *ICAPS*, pages 333–341.

[8] Cayton, L. (2008). Fast nearest neighbor retrieval for bregman divergences. In *Proceedings of the 25th international conference on Machine learning*, pages 112–119.

[9] Chandola, V., Banerjee, A., and Kumar, V. (2009). Anomaly detection: A survey. *ACM computing surveys (CSUR)*, 41(3):15.

[10] Crook, P., Hayes, G., et al. (2001). A robot implementation of a biologically inspired method for novelty detection. In *Proceedings of the Towards Intelligent Mobile Robots Conference*.

[11] Crook, P. A., Marsland, S., Hayes, G., and Nehmzow, U. (2002). A tale of two filters-on-line novelty detection. In *Proceedings 2002 IEEE International Conference on Robotics and Automation (Cat. No. 02CH37292)*, volume 4, pages 3894–3899. IEEE.

[12] Doherty, P., Kvarnström, J., and Heintz, F. (2009). A temporal logic-based planning and execution monitoring framework for unmanned aircraft systems. *Autonomous Agents and Multi-Agent Systems*, 19(3):332–377.

[13] Enrico Scala (n.d.). The enhsp planning system. https://sites.google.com/view/enhsp/. Accessed: 2022-05-06.

[14] Fox, M. and Long, D. (2003). Pddl2. 1: An extension to pddl for expressing temporal planning domains. *Journal of artificial intelligence research*, 20:61–124.

[15] Gil, Y. (1992). Acquiring domain knowledge for planning by experimentation. Technical report, CARNEGIE-MELLON UNIV PITTSBURGH PA DEPT OF COMPUTER SCIENCE.

[16] Glover, S. (2004). Separate visual representations in the planning and control of action. *Behavioral and brain sciences*, 27(1):3–24.

[17] Hadeel, J. (2022). *Improving Task Planning Knowledge Robustness for Autonomous Robots*. PhD thesis, University of Huddersfield.

[18] Häussermann, K., Zweigle, O., and Levi, P. (2015). A novel framework for anomaly detection of robot behaviors. *Journal of Intelligent & Robotic Systems*, 77(2):361–375.

[19] Hodge, V. and Austin, J. (2004). A survey of outlier detection methodologies. *Artificial intelligence review*, 22(2):85–126.

[20] Hornung, R., Urbanek, H., Klodmann, J., Osendorfer, C., and Van Der Smagt, P. (2014). Model-free robot anomaly detection. In *2014 IEEE/RSJ International Conference on Intelligent Robots and Systems*, pages 3676–3683. IEEE.

[21] Ingrand, F. and Ghallab, M. (2017). Deliberation for autonomous robots: A survey. *Artificial Intelligence*, 247:10–44.

[22] Jiménez, S., De la Rosa, T., Fernández, S., Fernández, F., and Borrajo, D. (2012). A review of machine learning for automated planning. *The*

*Knowledge Engineering Review*, 27(4):433–467.

[23] kCL planning (n.d.). Rosplan overview. *ROSPlan* (online). Accessed: 2021-01-03.

[24] Khalastchi, E., Kalech, M., Kaminka, G. A., and Lin, R. (2015). Online data-driven anomaly detection in autonomous robots. *Knowledge and Information Systems*, 43(3):657–688.

[25] Kirsch, A. and Beetz, M. (2007). Training on the job—collecting experience with hierarchical hybrid automata. In *Annual Conference on Artificial Intelligence*, pages 473–476. Springer.

[26] Krajnik, T., Fentanes, J. P., Cielniak, G., Dondrup, C., and Duckett, T. (2014). Spectral analysis for long-term robotic mapping. In *2014 IEEE International Conference on Robotics and Automation (ICRA)*, pages 3706–3711. IEEE.

[27] Kunze, L., Hawes, N., Duckett, T., Hanheide, M., and Krajník, T. (2018). Artificial intelligence for long-term robot autonomy: A survey. *IEEE Robotics and Automation Letters*, 3(4):4023–4030.

[28] Leonetti, M. and Iocchi, L. (2010). Learnpnp: A tool for learning agent behaviors. In *Robot Soccer World Cup*, pages 418–429. Springer.

[29] Lindsay, A., Franco, S., Reba, R., and McCluskey, T. L. (2020). Refining process descriptions from execution data in hybrid planning domain models. In *Proceedings of the International Conference on Automated Planning and Scheduling*, volume 30, pages 469–477.

[30] Louadah, H., Papadakis, E., McCluskey, L., Tucker, G., Hughes, P., and Bevan, A. (2021). Translating ontological knowledge to PDDL to do planning in train depot management operations. In *Proceedings of PlanSIG*. 36th Workshop of the UK Planning and Scheduling Special Interest Group, PlanSIG 2021.

[31] McDermott, D., Ghallab, M., Howe, A., Knoblock, C., Ram, A., Veloso, M., Weld, D., and Wilkins, D. (1998). Pddl-the planning domain definition language. *ROS Wiki*.

[32] Müller, A. and Beetz, M. (2006). Designing and implementing a plan library for a simulated household robot. In *Cognitive Robotics: Papers from the AAAI Workshop, Technical Report WS-06-03*, pages 119–128.

[33] Munzer, T., Toussaint, M., and Lopes, M. (2018). Efficient behavior learning in human–robot collaboration. *Autonomous Robots*, 42(5):1103–1115.

[34] Murphy, K. P. (2012). *Machine learning: a probabilistic perspective*. MIT press.

[35] Næs, T. and Risvik, E. (1996). *Multivariate analysis of data in sensory science*. Elsevier.

[36] Park, D., Hoshi, Y., and Kemp, C. C. (2018). A multimodal anomaly detector for robot-assisted feeding using an lstm-based variational autoencoder. *IEEE Robotics and Automation Letters*, 3(3):1544–1551.

[37] Schmidt, R. A. (1975). A schema theory of discrete motor skill learning. *Psychological review*, 82(4):225.

[38] Shen, W.-M. (1993). Discovery as autonomous learning from the environment. *Machine Learning*, 12(1):143–165.

[39] Smith, M. R. and Martinez, T. (2011). Improving classification accuracy by identifying and removing instances that should be misclassified. In *The 2011 International Joint Conference on Neural Networks*, pages 2690–2697. IEEE.

[40] Stulp, F. and Beetz, M. (2008). Refining the execution of abstract actions with learned action models. *Journal of Artificial Intelligence Research*, 32:487–523.

[41] Tan, P.-N., Steinbach, M., and Kumar, V. (2016). *Introduction to data mining*. Pearson Education India.

[42] Verma, V., Estlin, T., Jónsson, A., Pasareanu, C., Simmons, R., and Tso, K. (2005). Plan execution interchange language (plexil) for executable plans and command sequences. In *International symposium on artificial intelligence, robotics and automation in space (iSAIRAS)*. Citeseer.

[43] Wang, X. (1996). *Learning planning operators by observation and practice*. PhD thesis, Carnegie Mellon University.

[44] Weber, B. G., Mateas, M., and Jhala, A. (2012). Learning from demonstration for goal-driven autonomy. In *AAAI*.

[45] Zhang, L., Zhang, H., Yang, H., Bian, G.-B., and Wu, W. (2019). Multi-target detection and grasping control for humanoid robot nao. *International Journal of Adaptive Control and Signal Processing*, 33(7):1225–1237.

[46] Zimek, A. and Schubert, E. (2017). *Outlier Detection*, pages 1–5. Springer New York, New York, NY.

[47] Ziparo, V. A., Iocchi, L., Lima, P. U., Nardi, D., and Palamara, P. F. (2011). Petri net plans. *Autonomous Agents and Multi-Agent Systems*, 23(3):344–383.